\pdfoutput=1
\def\year{2019}\relax
\documentclass[letterpaper]{article} 
\usepackage{aaai19}  
\usepackage{times}  
\usepackage{helvet}  
\usepackage{courier}  
\usepackage{url}  
\usepackage{graphicx}  
\usepackage{multirow}
\usepackage{graphicx}
\usepackage{amsmath,amssymb,bm}

\begin{document}
%
\title{Synergistic Image and Feature Adaptation: Towards Cross-Modality\\ Domain Adaptation for Medical Image Segmentation}
\author{Cheng Chen$^{1}$, Qi Dou$^{1}$, Hao Chen$^{1,2}$, Jing Qin$^{3}$, and Pheng-Ann Heng$^{1,4}$\\
	$^1$ Department of Computer Science and Engineering, The Chinese University of Hong Kong\\
	$^2$ Imsight Medical Technology Co., Ltd., China\\
	$^3$ Centre for Smart Health, School of Nursing, The Hong Kong Polytechnic University,\\
	$^4$ Guangdong Provincial Key Laboratory of Computer Vision and Virtual Reality Technology, SIAT, China\\
	\{cchen, qdou, hchen, pheng\}@cse.cuhk.edu.hk, ~~~~harry.qin@polyu.edu.hk
}

\maketitle
\begin{abstract}
This paper presents a novel unsupervised domain adaptation framework, called \textit{Synergistic Image and Feature Adaptation (SIFA)}, to effectively tackle the problem of domain shift.
Domain adaptation has become an important and hot topic in recent studies on deep learning, aiming to recover performance degradation when applying the neural networks to new testing domains.
Our proposed SIFA is an elegant learning diagram which presents synergistic fusion of adaptations from both image and feature perspectives.
In particular, we simultaneously transform the appearance of images across domains and enhance domain-invariance of the extracted features towards the segmentation task. 
The feature encoder layers are shared by both perspectives to grasp their mutual benefits during the end-to-end learning procedure.
Without using any annotation from the target domain, the learning of our unified model is guided by adversarial losses, with multiple discriminators employed from various aspects.
We have extensively validated our method with a challenging application of cross-modality medical image segmentation of cardiac structures.
Experimental results demonstrate that our SIFA model recovers the degraded performance from 17.2\% to 73.0\%, and outperforms the state-of-the-art methods by a significant margin.
\end{abstract}

\section{Introduction}
\begin{figure}
	\includegraphics[width=0.47\textwidth]{{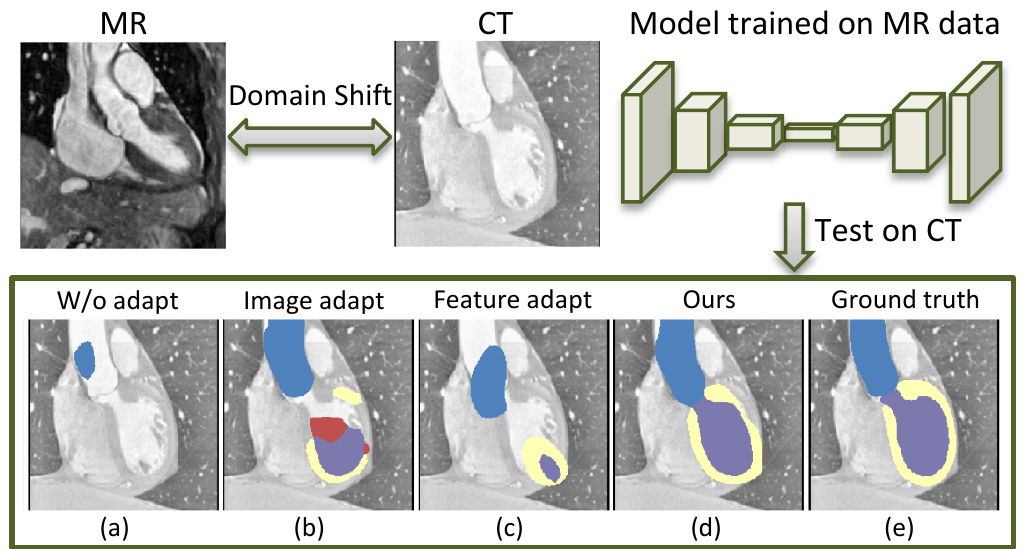}}
	\caption{Illustration of addressing the severe cross-modality domain shift of medical images from different perspectives. The segmentation results of the CT images with the DCNN trained on MR data are shown in the bottom: a) without any adaptation; b) with pure image adaptation; c) with pure feature adaptation; d) our proposed synergistic image and feature adaptations; e) the ground truth.}
\end{figure}

Deep convolutional neural networks (DCNNs) have made great breakthroughs in various challenging while crucial vision tasks~\cite{long2015fully,he2016deep}.
As investigations of DCNNs moving on, recent studies have frequently pointed out the problem of performance degradation when encountering domain shift,
i.e., attempting to apply the learned models on testing data (target domain) that have different distributions from the training data (source domain).
In medical image computing, which is an important area to apply AI for healthcare, the situation of heterogeneous domain shift is even more natural and severe, given the various imaging modalities with different physical principles.

For example, as shown in Fig.~1, the cardiac areas present significantly different visual appearance when viewed from different modalities of medical images, such as the magnetic resonance (MR) imaging and computed tomography (CT).
Unsurprisingly, the DCNNs trained on MR data completely fail when being tested on CT images. 
To recover model performance, an easy way is to re-train or fine-tune models with additional labeled data from the target domain~\cite{van2015transfer,ghafoorian2017transfer}. However, annotating data for every new domain is obviously and prohibitively expensive, especially in medical area that requires expertise.

To tackle this problem, unsupervised domain adaptation has been intensively studied to enable DCNNs to achieve competitive performance on unlabeled target data, only with annotations from the source domain.
Prior works have treated domain shift mainly from two directions. 
One stream is the \textit{image adaptation}, by aligning the image appearance between domains with the pixel-to-pixel transformation. 
In this way, the domain shift is addressed at input level to DCNNs.
To preserve pixel-level contents in original images, the adaptation process is usually guided by a cycle-consistency constraint~\cite{DBLP:conf/iccv/ZhuPIE17,hoffman2017cycada}. 
Typically, the transformed source-like images can be directly tested by pre-trained source models~\cite{russo2017source,zhang2018task}; alternatively,
the generated target-like images can be used to train models in target domain~\cite{bousmalis2017unsupervised,zhao2018supervised}.
Although the synthesis images still cannot perfectly mimic the appearance of real images, the image adaptation process brings accurate pixel-wise predictions on target images, as shown in Fig.~1.

The other stream for unsupervised domain adaptation follows the \textit{feature adaptation}, which aims to extract domain-invariant features with DCNNs, regardless of the appearance difference between input domains.
Most methods within this stream discriminate feature distributions of source/target domains in an adversarial learning scenario~\cite{ganin2016domain,tzeng2017adversarial,DBLP:conf/ijcai/DouOCCH18}.
Furthermore, considering the high-dimensions of plain feature spaces, some recent works connected the discriminator to more compact spaces.
For examples, Tsai et al. inputs segmentation masks to the discriminator, so that the supervision arises from a semantic prediction space~\cite{tsai2018learning}.
Sankaranarayanan et al. reconstructs the features into images and put a discriminator in the reconstructed image space~\cite{sankaranarayanan2018learning}.
Although the adversarial discriminators implicitly enhance domain invariance of features extracted by DCNNs, the adaptation process can output results with proper and smooth shape geometry.

Being aware that the image adaptation and feature adaptation address domain shift from complementary perspectives, we recognize that the two adaptation procedures can be performed together within one unified framework.
With image transformation, the source images are transformed towards the appearance of target domain; afterwards, the remaining gap between the synthesis target-like images and real target images can be further addressed using the feature adaptation.
Sharing this spirit, several very recent works have presented promising attempts using image and feature adaptations altogether~\cite{hoffman2017cycada,zhang2018fully}. 
However, these existing methods conduct the two perspectives of adaptations sequentially, without leveraging mutual interactions and benefits.
Surely, there still remains extensive space for synergistic merge of image and feature adaptations, to elegantly overcome hurdle of domain shift when generalizing DCNNs to new domains with zero extra annotation cost.

In this paper, we propose a novel unsupervised domain adaptation framework, called \textit{Synergistic Image and Feature Adaptation (SIFA)}, and successfully apply it to adaptation of cross-modality medical image segmentation under severe domain shift. 
Our designed SIFA presents an elegant learning diagram which enables synergistic fusion of adaptations from both image and feature perspectives.
More specifically, we transform the labeled source images into the appearance of images drawn from the target domain, by using generative adversarial networks with cycle-consistency constraint.
When using the synthesis target-like images to train a segmentation model, we further integrate feature adaptation to combat the remaining domain shift.
Here, we use two discriminators, respectively connecting the semantic segmentation predictions and generated source-like images, to differentiate whether obtained from synthesis or real target images.
Most importantly, in our designed SIFA framework, we share the feature encoder, such that it can simultaneously transform image appearance and extract domain-invariant representations for the segmentation task. The entire domain adaptation framework is unified and both image and feature adaptations are seamlessly integrated into an end-to-end learning diagram. 
The major contributions of this paper are as follows: 

\begin{itemize}
	
	\item We present the SIFA, a novel unsupervised domain adaptation framework, that exploits synergistic image and feature adaptations to tackle domain shift via complementary perspectives.
	
	\item We enhance feature adaptation by using discriminators in two aspects, i.e., semantic prediction space and generated image space.
	Both compact spaces help to further enhance domain-invariance of the extracted features.
	
	\item We validate the effectiveness of our SIFA on the challenging task of cross-modality cardiac structure segmentation.
	Our approach recovers the performance degradation from 17.2\% to 73.0\%, and outperforms the state-of-the-art methods by a significant margin. The code is available at \url{https://github.com/cchen-cc/SIFA}.

\end{itemize}

\section{Related Work}
\begin{figure*}
	\centering
	\includegraphics[width=0.92\textwidth]{{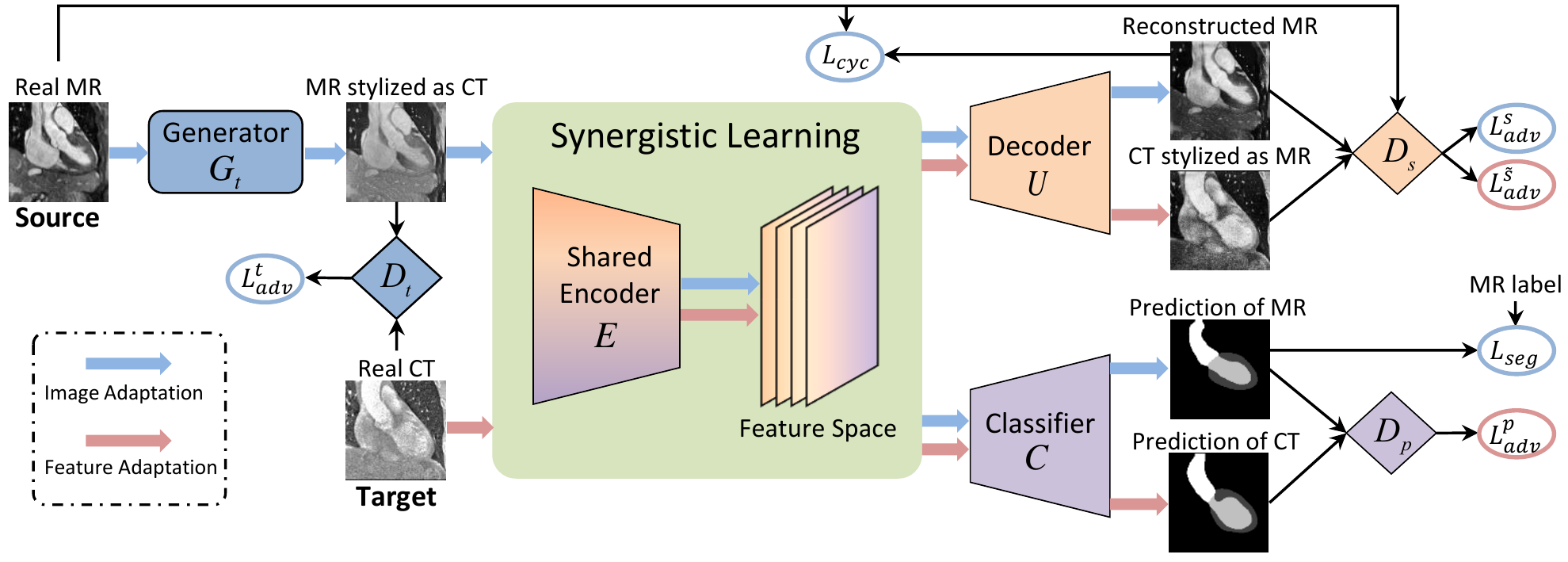}}
	\caption{Overview of our unsupervised domain adaptation framework. 
		The generator $G_t$ serves the source-to-target image transformation. The encoder $E$ and decoder $U$ form the reverse transformation, where the encoder $E$ is also connected with a classifier $C$ for image segmentation. The discriminators $\{D_t,D_s,D_p\}$ differentiate their inputs accordingly to derive adversarial losses. 
		The blue and red arrows indicate the data flows for the image adaptation and feature adaptation respectively. The reverse cycle-consistency is omitted in this figure for ease of illustration.}
\end{figure*}

Addressing performance degradation of DCNNs under domain shift has been a highly active and fruitful research field in recent investigations of deep learning.
A plentiful of adaptive methods have been proposed from different perspectives, including the image-level adaptation, feature-level adaptation and their mixtures. In this section, we overview the progress and state-of-the-art approaches along these streams, with a particular focus on unsupervised domain adaptation in image processing field. Studies on both natural and medical images are covered.

With a gratitude to generative adversarial network~\cite{DBLP:conf/nips/GoodfellowPMXWOCB14},
image-level adaptation methods have been developed to tap domain shift at the input level to DCNNs. 
Some methods first trained a DCNN in source domain, and then transformed the target images into source-like ones, such that can be tested using the pre-trained source model~\cite{russo2017source,zhang2018task,chen2018semantic}. 
Inversely, other methods tried to transform the source images into the appearance of target images~\cite{bousmalis2017unsupervised,shrivastava2017learning,hoffman2017cycada}.
The transformed target-like images are then used to train a task model which could perform well in target domain. 
This has also been used in medical eye retinal fundus image analysis~\cite{zhao2018supervised}.
With the wide success of CycleGAN in unpaired image-to-image transformation, many previous image adaptation works were based on modified CycleGAN with applications in both natural datasets~\cite{russo2017source,hoffman2017cycada} and medical image segmentation~\cite{huo2018adversarial,zhang2018task,chen2018semantic}.

Meanwhile, approaches for feature-level adaptation have also been investigated, aiming to reduce domain shift by extracting domain-invariant features in the DCNNs.
Pioneer works tried to minimize the distance between domain statistics, such as the maximum mean distance~\cite{DBLP:conf/icml/LongC0J15} and the layer activation correlation~\cite{sun2016deep}.
Later, representative methods of DANN~\cite{ganin2016domain} and ADDA~\cite{tzeng2017adversarial} advanced feature adaptation via adversarial learning, by using a discriminator to differentiate the feature space across domains.
Effectiveness of this strategy has also been validated in medical applications of segmenting brain lesions~\cite{kamnitsas2017unsupervised} and 
cardiac structures~\cite{DBLP:conf/ijcai/DouOCCH18,joyce2018deep}.
Recent studies proposed to project the high-dimensional feature space to other compact spaces, such as the semantic prediction space~\cite{tsai2018learning} or the image space~\cite{sankaranarayanan2018learning}, and a discriminator operated in the compact spaces to derive adversarial losses for more effective feature alignment.

The image and feature adaptations address domain shift from different perspectives to the DCNNs, which are in fact complementary to each other. 
Combining these two adaptive strategies to achieve a stronger domain adaption technique is under explorable progress.
As the state-of-the-art methods for semantic segmentation adaptation methods,
the CyCADA~\cite{hoffman2017cycada} and Zhang et al.~\cite{zhang2018fully} achieved leading performance in adaptation between synthetic to real world driving scene domains. However, their image and feature adaptations are sequentially connected and trained in stages without interactions.

Considering the severe domain shift in cross-modality medical images, feature adaptation or image adaptation alone may not be sufficient in this challenging task while the simultaneous adaptations from the two perspectives have not been fully explored yet. 
To tackle the challenging cross-modality adaptation for segmentation task, we propose to synergistically merge the two adaptive processes in a unified network to fully exploit their mutual benefits towards unsupervised domain adaptation.

\section{Methods}
An overview of our proposed method for unsupervised domain adaptation in medical image segmentation is shown in Fig.~2. 
We propose synergistic image and feature adaptations with a novel learning diagram to effectively narrow the performance gap due to domain shift.
The two perspectives of adaptations are seamlessly integrated into a unified model, and hence, both aspects can mutually benefit each other during the end-to-end training procedure.

\subsection{Image Adaptation for Appearance Alignment}
First, with a set of labeled samples $\{x_i^s,y_i^s\}_{i=1}^N$ from the source domain $X^s$, as well as unlabeled samples $\{x_j^t\}_{j=1}^M$ from the target domain $X^t$,
we aim to transform the source images $x^s$ towards the appearance of target ones $x^t$, which hold different visual appearance due to domain shift.
The obtained transformed image looks as if drawn from the target domain, while the original contents with structural semantics remain unaffected. 
Briefly speaking, this module narrows the domain shift between the source and target domains by aligning image appearance.

In practice, we use generative adversarial networks, which have made a wide success for pixel-to-pixel image transformation, by building a generator $G_t$ and a discriminator $D_t$.
The generator aims to transform the source images to target-like ones $G_t(x^s)\! = \! x^{s\to t}$.
The discriminator competes with the generator to correctly differentiate the fake transformed image $x^{s\to t}$ and the real target image $x^t$.
Therefore, in the target domain, the $G_t$ and $D_t$ form a minimax two-player game and are optimized via the adversarial learning:
\begin{equation}
\small
\begin{split}
\mathcal{L}_{\textit{adv}}^t(G_t,D_t) = &~\mathbb{E}_{x^{t}\sim X^{t}}[\text{log}D_t(x^{t})]+\\
&~\mathbb{E}_{x^{s}\sim X^{s}}[\text{log}(1-D_t(G_t(x^{s})))],
\end{split}
\end{equation}
where the discriminator tries to maximize this objective to distinguish between $G_t(x^s)\! = \! x^{s\to t}$ and $x^t$, and meanwhile, the generator needs to minimize this objective to transform $x^s$ into realistic target-like images.

To preserve original contents in the transformed images, a reverse generator is usually used to impose the cycle consistency~\cite{DBLP:conf/iccv/ZhuPIE17}.
As shown in Fig.~2, the encoder $E$ and upsampling decoder $U$ form the reverse target-to-source generator $G_s \! = \!E \circ U $ to reconstruct the $x^{s\to t}$ back to the source domain, and a discriminator $D_s$ operates in the source domain.
This pair of source $\{G_s, D_s\}$ are trained in the same manner as $\{G_t, D_t\}$ with the adversarial loss $\mathcal{L}_{\textit{adv}}^s$.
Then the pixel-wise cycle-consistency loss $\mathcal{L}_{\textit{cyc}}$ is used to encourage $U(E(G_t(x^s)))\approx x^s$ and $G_t(U(E(x^t)))\approx x^t$ for recovering the original image:
\begin{equation}
\small
\begin{split}
\mathcal{L}_{\textit{cyc}}(G_t,E,U) = & ~ \mathbb{E}_{x^{s}\sim X^s}||U(E(G_t(x^s)))-x^s||_1+\\
&~\mathbb{E}_{x^{t}\sim X^t}||G_t(U(E(x^t)))-x^t||_1.
\end{split}
\end{equation}

With the adversarial loss and cycle-consistency loss, the image adaptation transforms the source images $x^s$ into target-like images $x^{s\to t}$ with semantic contents preserved.
Ideally, this pixel-to-pixel transformation could bring $x^{s\to t}$ into the data distribution of target domain, such that these synthesis images can be used to train a segmentation network for the target domain.

Specifically, after extracting features from the adapted image $x^{s\to t}$, the feature maps $E(x^{s\to t})$ are forwarded to a classifier $C$ for predicting segmentation masks. In other words, the composition of $E \circ C$ serves as the segmentation network for the target domain.
This part is trained using the sample pairs of $\{x^{s\to t}, y^s\}$ by minimizing a hybrid loss $\mathcal{L}_{\textit{seg}}$. Formally, denoting the segmentation prediction for $x^{s\to t}$ by $\hat{y}^{s\to t} \! = \! C(E(x^{s\to t}))$, the segmentation loss is defined as:
\begin{equation}
\small
\mathcal{L}_{\textit{seg}}(E,C)=H(y^s,\hat{y}^{s\to t})+\alpha \cdot \textit{Dice}(y^s,\hat{y}^{s\to t}),
\end{equation}
where the first term represents cross-entropy loss, the second term is the Dice loss, and $\alpha$ is the trade-off hyper-parameter balancing them.
The hybrid loss function is designed to meet the class imbalance in medical image segmentation.

\subsection{Feature Adaptation for Domain Invariance}

In above image adaptation, training a segmentation network with the transformed target-like images can already get appealing performance on target data.
Unfortunately, when domain shift is severe, such as for cross-modality medical images, it is still insufficient to achieve desired domain adaptation results. To this end, we further impose additional discriminators to contribute from the perspective of feature adaptation, attempting to bridge the remaining domain gap between the synthesis target images and real target images.

To make the extracted features domain-invariant, the most common way is using adversarial learning directly in feature space, such that a discriminator fails to differentiate which features come from which domain. However, a feature space is with high-dimension, and hence difficult to be directly aligned. Instead, we choose to enhance the domain-invariance of feature distributions by using adversarial learning via two compact lower-dimensional spaces.
Specifically, we inject adversarial losses via the semantic prediction space and the generated image space.

As shown in Fig.~2, for prediction of segmentation masks from $\{E,C\}$, we construct the discriminator $D_p$ to classify the outputs corresponding to $x^{s\to t}$ or $x^t$. 
The semantic prediction space represents the information of human-body anatomical structures, which should be consistent across different imaging modalities. If the features extracted from $x^{s\to t}$ are aligned with that from $x^t$, the discriminator $D_p$ would fail in differentiating their corresponding segmentation masks, as the anatomical shapes are consistent. Otherwise, the adversarial gradients are back-propagated to the feature extractor $E$, so as to minimize the distance between the feature distributions from $x^{s\to t}$ and $x^t$. 
The adversarial loss from semantic-level supervision for the feature adaptation is: 
\begin{equation}
\small
\begin{split}
\mathcal{L}_{\textit{adv}}^{p}(E,C,D_p)= &\mathbb{E}_{x^{s\to t}\sim X^{s\to t}}[\text{log}~D_p(C(E(x^{s\to t})))]+\\
&\mathbb{E}_{x^{t}\sim X^t}[\text{log}(1-D_p(C(E(x^{t}))))].
\end{split}
\end{equation}

For generated source-like images from $\{E, U\}$, we add an auxiliary task to the source discriminator $D_s$ to differentiate whether the generated images are transformed from real target images $x^t$ or reconstructed from $x^{s\to t}$. 
If the discriminator $D_s$ succeeded in classifying the domain of generated images, it means that the extracted features still contain domain characteristics. To make the features domain-invariant, the following adversarial loss is employed to supervise the feature extraction process:
\begin{equation}
\small
\begin{split}
\mathcal{L}_{\textit{adv}}^{\tilde{s}}(E,D_s) = & ~\mathbb{E}_{x^{s\to t}\sim X^{s\to t}}[\text{log} D_s(U(E(x^{s\to t})))] + \\
&~\mathbb{E}_{x^{t}\sim X^t}[\text{log}(1-D_s(U(E(x^{t}))))].
\end{split}
\end{equation}

It is noted that the $E$ is encouraged to extract features with domain-invariance by connecting discriminator from two aspects, i.e.,
segmentation predictions (high-level semantics) and generated source-like images (low-level appearance).
By adversarial learning from these lower-dimensional compact spaces, the domain gap between synthesis target images $x^{s\to t}$ and real target images $x^t$ can be effectively addressed.

\subsection{Synergistic Learning Diagram}

Importantly, a key characteristic in our proposed synergistic learning diagram is to share the feature encoder $E$ between both image and feature adaptations.
More specifically, the $E$ is optimized with the adversarial loss $\mathcal{L}_{\textit{adv}}^s$ and cycle-consistency loss $\mathcal{L}_{\textit{cyc}}$ via the image adaptation perspective. It also collects gradients back-propagated from the discriminators $\{D_p, D_s\}$ towards feature adaptation.
In these regards, the feature encoder is fitted inside a multi-task learning scenario, such that, it is able to present generic and robust representations useful for multiple purposes.
In turn, the different tasks bring complementary inductive bias to the encoder parameters, i.e., either emphasizing pixel-wise cyclic reconstruction or focusing on structural semantics.
This can also contribute to alleviate the over-fitting problem with limited medical datasets when training such a complicated model.

With the encoder enabling seamless integration of the image and feature adaptations, we can train the unified framework in an end-to-end manner.
At each training iteration, all the modules are sequentially updated in the following order: $G_t$ $\rightarrow$ $D_t$ $\rightarrow$ $E$ $\rightarrow$ $C$ $\rightarrow$ $U$ $\rightarrow$ $D_s$ $\rightarrow$ $D_p$. 
Specifically, The generator $G_t$ is updated first to obtain the transformed target-like images. Then the discriminator $D_t$ is updated to differentiate the target-like images $x^{s\to t}$ and the real target images $x^t$. Next, the encoder $E$ is updated for feature extraction from $x^{s\to t}$ and $x^t$, followed by the updating of classifier $C$ and decoder $U$ to map the extracted features to the segmentation predictions and generated source-like images. Finally, the discriminator $D_s$ and $D_p$ are updated to classify the domain of their inputs to enhance feature-invariance.
The overall objective for our framework is as follows:
\begin{equation}
\small
\begin{split}
\mathcal{L} =&~\mathcal{L}_{\textit{adv}}^{t}(G_t,D_t)+\lambda_{\textit{adv}}^{s}\mathcal{L}_{\textit{adv}}^{s}(E,U,D_s)~+~\\    &~\lambda_{\textit{cyc}}\mathcal{L}_{\textit{cyc}}(G_t,E,U) +\lambda_{\textit{seg}}\mathcal{L}_{\textit{seg}}(E,C)~+~\\
&~\lambda_{\textit{adv}}^{p}\mathcal{L}_{\textit{adv}}^{p}(E,C,D_p)+\lambda_{\textit{adv}}^{\tilde{s}}\mathcal{L}_{\textit{adv}}^{\tilde{s}}(E,D_s)\\
\end{split}
\end{equation}
where the $\{\lambda_{\textit{adv}}^{s},\lambda_{\textit{cyc}},\lambda_{\textit{seg}},\lambda_{\textit{adv}}^{p},\lambda_{\textit{adv}}^{\tilde{s}}\}$ are trade-off parameters adjusting the importance of each component. 

For training practice, when updating with the adversarial learning losses, we used the Adam optimizer with a learning rate of $2\!\times\!10^{-4}$.
For segmentation task, the Adam optimizer was parameterized with an initial learning rate of $1\!\times\!10^{-3}$ and a stepped decay rate of 0.9 every 2 epochs. 

During the testing procedure, when an image from the target domain arrives, this $x^t$ is forwarded into the encoder $E$, followed by applying the classifier $C$. In this way, the semantic segmentation result is obtained by $C(E(x^t))$, using the domain adaptation framework which is learned without need of any target domain annotations.

\subsection{Network Configurations of the Modules}

In this section, we describe the detailed network configurations of every module in the proposed framework.
Residual connections are widely used to ease the gradients flow inside our complicated model.
We also actively borrow the previous successful experiences of training generative adversarial networks, as reported in the references.

The layer configuration of the target generator $G_t$ follow the practice of CycleGAN \cite{DBLP:conf/iccv/ZhuPIE17}. It consists of 3 convolutional layers, 9 residual blocks, and 2 deconvolutional layers, finally using one convolutional layer to get the generated images. For the source decoder $U$, we construct it with 1 convolutional layer, 4 residual blocks, and 3 deconvolutional layers, finally also followed by one convolutional output layer. 
For all the three discriminators $\{D_t, D_s, D_p\}$, we follow the configuration of PatchGAN \cite{DBLP:conf/cvpr/IsolaZZE17}, by differentiating $70 \! \times \! 70$ patches. The networks consist of 5 convolutional layers with kernels as size of $4 \! \times \! 4$ and stride of 2, except for the last two layers, which use convolution stride of 1. The numbers of feature maps are $\{64,128,256,512,1\}$ for each layer, respectively. At the first four layers, each convolutional layer is followed by an instance normalization and a leaky ReLU parameterized with $0.2$.

The encoder $E$ uses residual connections and dilated convolutions ($\textit{dilation rate}\! = \! 2$) to enlarge the size of receptive field while preserving the spatial resolution for dense predictions~\cite{yu2017dilated}. 
Let $\{\text{C}k,\text{R}k,\text{D}k\}$ denote a convolutional layer, a residual block and a dilated residual block with $k$ channels, respectively.
The $\text{M}$ represents the max-pooling layer with a stride of 2.
Our encoder module is deep by stacking layers of
$\{\text{C16},\text{R16},\text{M},\text{R32},\text{M},2\!\times\!\text{R64},\text{M},2\!\times\!\text{R128},4\!\times\!\text{R256},2\!\times\!\text{R512},2\!\times\!\text{D512},2\!\times\!\text{C512}\}$. 
Each convolution operation is connected to a batch normalization layer and ReLU activation.
The classifier $C$ is a $1\!\times\!1$ convolutional layer followed by an upsampling layer to recover the resolution of segmentation predictions to original image size.   

\section{Experimental Results}
\begin{figure*}[h!]
	\centering
	\includegraphics[width=1\textwidth]{{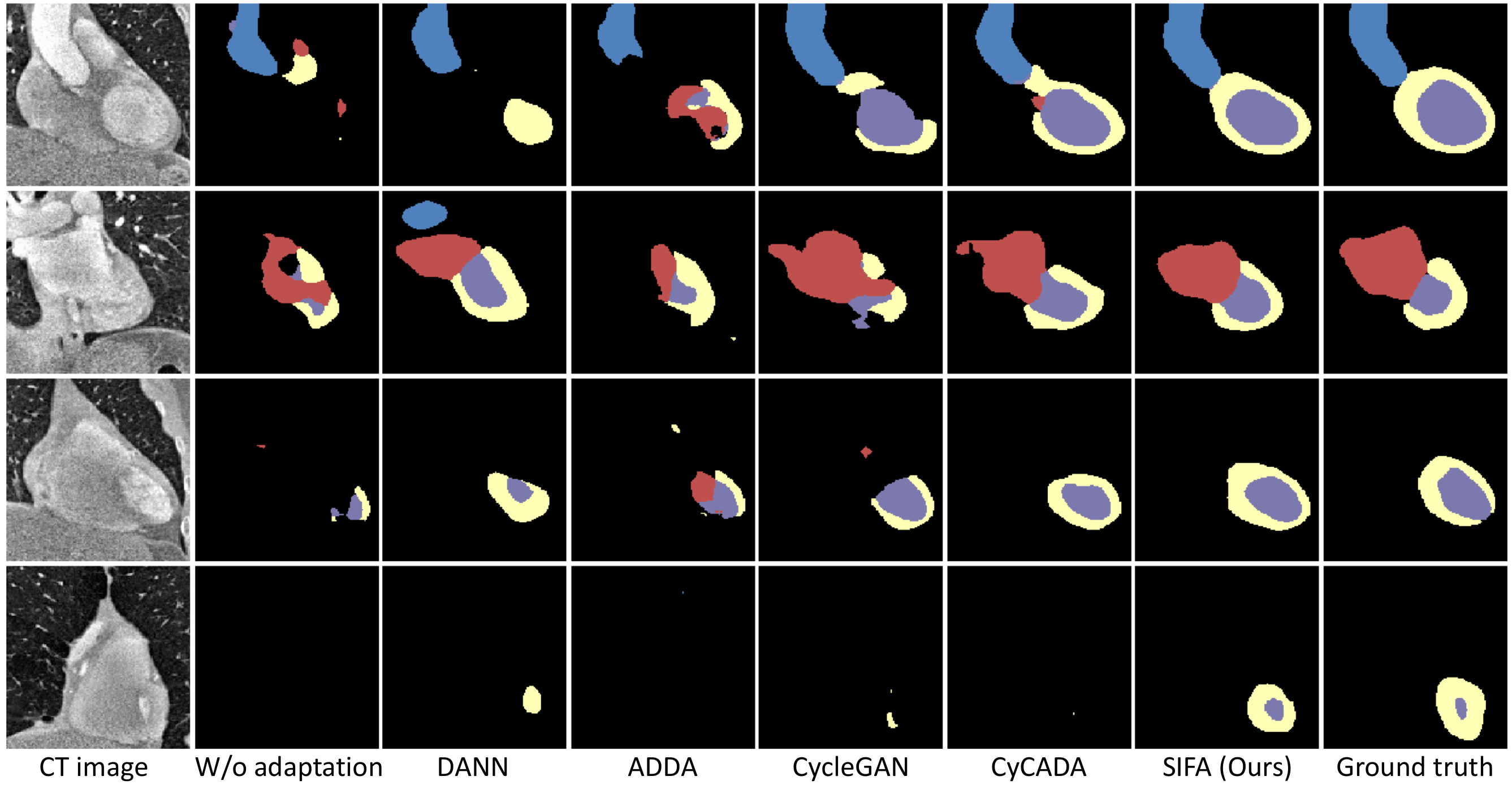}}
	\caption{Visual comparison of segmentation results produced by different methods. From left to right are the raw CT images (1st column), "W/o Adaptation" lower bound (2nd column), results of other unsupervised domain adaptation methods (3rd-6th column), results of our SIFA network (7th column), and ground truth (last column). The cardiac structures of AA, LAC, LVC, and MYO are indicated in blue, red, purple, and yellow color respectively. Each row corresponds to one example.
	}
\end{figure*}
\begin{table*}[h!]
	\caption{Performance comparison between our method and other state-of-the-art unsupervised domain adaptation methods for the task of cardiac cross-modality segmentation. We report the Dice and ASD value for each cardiac structure and the average of the four structures. (Note: - means that the results are not reported by that methods and N/A means that the ASD value cannot be calculated due to no prediction for that cardiac structure.)
	}
\centering
\begin{center}
	\resizebox{1\textwidth}{!}{%
		\begin{tabular}{c|cc|ccccc|ccccc}
			\hline
				
			\multirow{2}{*}{Methods} &\multicolumn{2}{c|}{Adaptation}&\multicolumn{5}{c|}{Dice}&\multicolumn{5}{c}{ASD}\\
			\cline{2-13}
			&Image  &Feature &AA &LAC &LVC &MYO &Average &AA &LAC &LVC &MYO &Average \\
			
			\hline
			\hline
			W/o adaptation & & &28.4 &27.7 &4.0 &8.7 &17.2 &20.6 &16.2 &N/A &48.4 &N/A\\
			
			\hline
			\hline
			DANN~\cite{ganin2016domain} & &\checkmark &39.0 &45.1 &28.3 &25.7 &34.5 &16.2 &9.2 &12.1 &10.1 &11.9 \\
			
			ADDA~\cite{tzeng2017adversarial} & &\checkmark &47.6 &60.9 &11.2 &29.2 &37.2 &13.8 &10.2 &N/A &13.4 &N/A \\
			
			CycleGAN~\cite{DBLP:conf/iccv/ZhuPIE17} &\checkmark & &73.8 &75.7 &52.3 &28.7 &57.6 &11.5 &13.6 &9.2 &8.8 &10.8\\
			
			CyCADA~\cite{hoffman2017cycada} &\checkmark &\checkmark &72.9 &\textbf{77.0} &62.4 &45.3 &64.4 &\textbf{9.6} &8.0 &9.6 &10.5 &9.4 \\
			
			\hline
			\hline
			Dou et al.~\cite{DBLP:conf/ijcai/DouOCCH18} & &\checkmark &74.8 &51.1 &57.2 &47.8 &57.7&27.5 &20.1 &29.5 &31.2&27.1 \\
			
			Joyce et al.~\cite{joyce2018deep} & &\checkmark &- &- &66 &44&- &-&-&-&-&-\\
			
			\hline
			\hline
			SIFA (Ours) &\checkmark &\checkmark &\textbf{81.1} &76.4 &\textbf{75.7} &\textbf{58.7}&\textbf{73.0} &10.6&\textbf{7.4}&\textbf{6.7}&\textbf{7.8}&\textbf{8.1}\\
			\hline
		\end{tabular}}
	\end{center}
\end{table*}
	
\subsection{Dataset and Evaluation Metrics}
We validated our proposed unsupervised domain adaptation method on the \textit{Multi-Modality Whole Heart Segmentation Challenge 2017} dataset for cardiac segmentation in MR and CT images~\cite{zhuang2016multi}. 
The dataset consists of unpaired 20 MR and 20 CT volumes collected at different clinical sites. 
The ground truth masks of cardiac structures are provided, including the ascending aorta (AA), the left atrium blood cavity (LAC), the left ventricle blood cavity (LVC), and the myocardium of the left ventricle (MYO).
We aim to adapt the segmentation network at the setting of cross-modality learning.

We employed the MR images as the source domain, and the CT images as the target domain.  
Each modality was randomly split with 80\% cases for training and 20\% cases for testing. The ground truth of CT images were used for evaluation only, without being presented to the network during training phase.
All the data were normalized as zero mean and unit variance. To train our model, we used the coronal view images slices, which were cropped into the size of $256\!\times\!256$ and augmented with rotation, scaling, and affine transformations to reduce over-fitting.

For evaluation, we employed two commonly-used metrics to quantitatively evaluate the segmentation performance, which have also been used in previous cross-modality domain adaptation works~\cite{DBLP:conf/ijcai/DouOCCH18,joyce2018deep}. One measurement is the Dice coefficient ([\%]), which calculates the volume overlap between the prediction mask and the ground truth. The other is the average surface distance ASD ([voxel]) to assess the model performance at boundaries and a lower ASD indicates the better segmentation results. 
	
\subsection{Comparison with the State-of-the-art Methods}

We compare our framework with six recent popular unsupervised domain adaptation methods including DANN~\cite{ganin2016domain}, ADDA~\cite{tzeng2017adversarial}, CycleGAN~\cite{DBLP:conf/iccv/ZhuPIE17}, CyCADA~\cite{hoffman2017cycada}, Dou et al.~\cite{DBLP:conf/ijcai/DouOCCH18}, and Joyce et al.~\cite{joyce2018deep}. 
Among them, The first four are proposed for natural datasets, and we either used public available code or re-implemented them for our cardiac segmentation dataset. The DANN and ADDA employ only feature adaptation, the CycleGAN adapts image appearance, and the CyCADA conducts both image and feature adaptations.
The last two methods are dedicated to adapt MR/CT cardiac segmentation networks in feature level using the same cross-modality dataset as ours, therefore, for which we directly reference the results from their papers. 
We also obtain the "W/o adaptation" lower bound by directly applying the model learned in MR source domain to test target CT images without using any domain adaptation method.

\begin{figure}[t!]
	\centering
	\includegraphics[width=0.45\textwidth]{{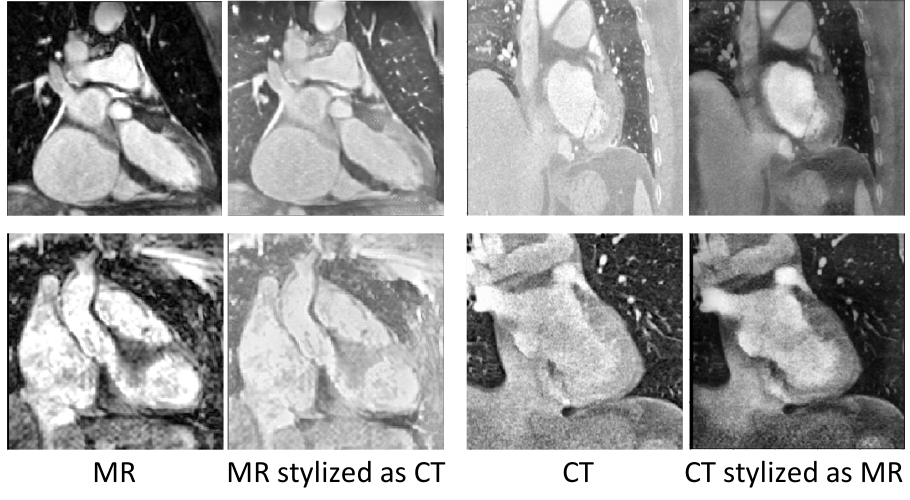}}
	\caption{Examples of image transformation between MR and CT images.}
\end{figure}
\begin{table}[h!]
	\centering
	\caption{Effectiveness of each key component in SIFA. "IA" denotes image adaptation; "FA-P" and "FA-I" respectively denote the feature adaptation in the semantic prediction space and the generated image space.}
	\begin{center}
		\resizebox{0.47\textwidth}{!}{%
			\begin{tabular}{ l|ccc|c  }
				\hline
				Methods &~~IA~~&$\mathcal{L}_{adv}^{p}$&$\mathcal{L}_{adv}^{\tilde{s}}$&Average Dice\\
				
				\hline
				\hline
				W/o adaptation &&&&17.2\\
				+~Image adaptation &\checkmark&&&58.0 \\
				+~FA-P&\checkmark&\checkmark&&65.7 \\
				+~FA-I &\checkmark&\checkmark&\checkmark&\textbf{73.0} \\	
				
				\hline
			\end{tabular}
		}
	\end{center}
\end{table}

Table 1 reports the comparison results, where we can see that our method significantly increased the segmentation performance over the "W/o adaptation" lower bound and outperformed previous methods by a large margin in terms of both Dice and ASD. 
Without domain adaptation, the model only obtained the average Dice of 17.2\% over the four cardiac structures, demonstrating the severe domain shift between MR and CT images. 
Remarkably, with our SIFA network, the average Dice was recovered to 73.0\% and the average ASD was reduced to 8.1.
We achieved over 80\% Dice score for the AA structure and over 70\% Dice score for the LAC and LVC.
Notably, compared with CyCADA, which also conducts both image and feature adaptations, our method achieved superior performance especially for the LVC and MYO structures, which have limited contrast in CT images. This demonstrates the effectiveness of our synergistic learning diagram, which unleashes the benefits from mutual conduction of image and feature alignments. 

Visual comparison results are further provided in Fig.~3. We can see that without adaptation, the network hardly outputs any correct prediction for the cardiac structures. 
By using feature adaptation (3rd and 4th columns) or image adaptation (5th column) alone, appreciable recovery in the segmentation prediction masks can be obtained, but the shape of predicted cardiac structures is quite cluttered and noisy. 
Only the two methods, CyCADA and our SIFA, which leverage both the feature and image adaptations, can generate semantically meaningful prediction for the four cardiac structures. Particularly, our SIFA network outperforms CyCADA especially for the segmentation of LVC and MYO. As can be seen in the last row in Fig.~3, the LVC and MYO structures have very limited intensity contrast with their surrounding tissues, but our method can make good predictions while all the other methods fail in this challenging case. 

\subsection{Effectiveness of Key Components}

We conduct ablation experiments to evaluate the effectiveness of each key component in our proposed synergistic learning framework of image and feature adaptations. The results are presented in Table 2.
Our baseline network uses image adaptation only, which is constructed by removing the feature adaptation adversarial loss $\mathcal{L}_{\textit{adv}}^{p}$ and $\mathcal{L}_{\textit{adv}}^{\tilde{s}}$ when training the network, i.e., removing the data flow of red arrows in Fig.~2. Compared with the "W/o adaptation" lower bound, our baseline network with pure image adaptation already achieved inspiring increase in segmentation accuracy with average Dice increased to 58.0\%. This reflects that with image transformation, the source images have been successfully brought closer to the target domain.
Fig.~4 shows four examples of image transformation from source to target domain and vice versa. 
As illustrated in the figure, the appearance of images is successfully adapted across domains while the semantic contents in original images are well-preserved. 

Next, we combine baseline image adaptation with one aspect of feature adaptation, i.e., adding the adversarial learning in the semantic prediction space, which corresponds to adding the discriminator guided by $\mathcal{L}_{\textit{adv}}^{p}$. 
The increased performance over the image adaptation baseline, from 58.0\% to 65.7\%, demonstrates that the image and feature adaptations are complementary to each other and can be jointly conducted to achieve better domain adaptation. 
Finally, further adding the feature adaptation by aligning generated source-like images with $\mathcal{L}_{\textit{adv}}^{\tilde{s}}$ completes our full SIFA network. This leads to further obvious improvement in the average Dice accuracy of segmentation results, indicating that the feature adaptation in these two compact spaces would inject effects from integral aspects to encourage feature invariance.

Fig.~5 shows the visual comparison results of our network with different components.
We can see that the segmentation results become increasingly accurate as more adaptation components being included. 
Our baseline network with image adaptation alone can correctly identify the cardiac structures, but the predicted shape is irregular and noisy. Adding the feature adaptation in the two lower-dimensional spaces further encourages the network to capture the proper shape of cardiac structures and produce clear predictions. Overall, our SIFA network synergistically merges different adaptation strategies to exploit their complementary contributions to unsupervised domain adaptation.

\begin{figure}[t!]
	\centering
	\includegraphics[width=0.45\textwidth]{{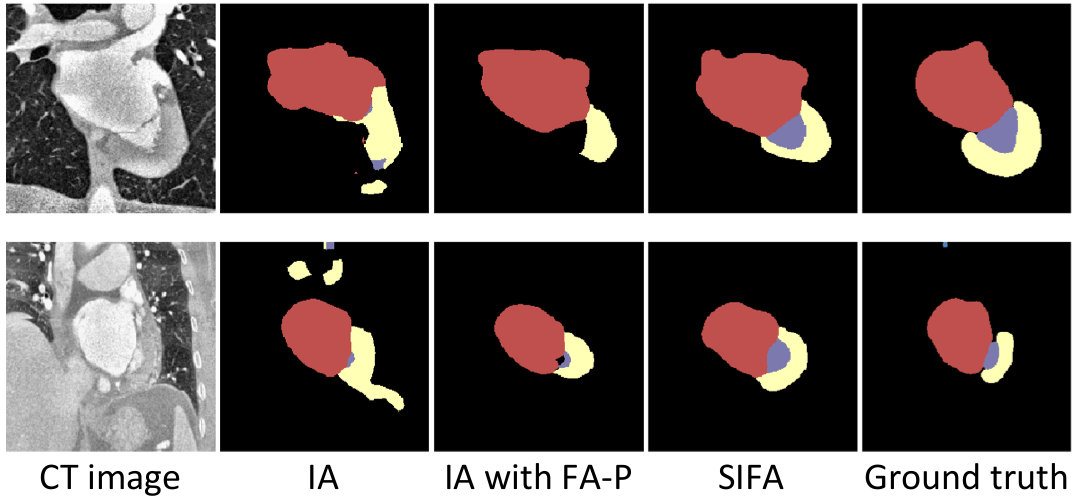}}
	\caption{Illustration of effectiveness of each key component in our method: "IA" denotes our network with image adaptation only; "IA with FA-P" denotes the combination of image adaptation and the feature adaptation in semantic prediction space; "SIFA" is our overall framework.}
\end{figure}
		
\section{Conclusion}
This paper proposes a novel approach SIFA for unsupervised domain adaptation of cross-modality medical image segmentation. 
Our SIFA network synergistically combines the image and feature adaptations to conduct image appearance transformation and domain-invariant feature learning simultaneously. 
The two adaptive perspectives are guided by the adversarial learning with partial parameter sharing to exploit their mutual benefits for reducing domain shift during the end-to-end training. 
We validate our method on unpaired MR to CT adaptation for cardiac segmentation by comparing it with various state-of-the-art methods. 
Experimental results demonstrate the superiority of our network over the others in terms of both the Dice and ASD value.
Our method is general and can be easily extended to other segmentation applications of unsupervised domain adaptation. 

\section{Acknowledgments}
This work was supported by a grant from 973 Program (Project No. 2015CB351706), a grant from Shenzhen Science and Technology Program (JCYJ20170413162256793), a grant from the Hong Kong Research Grants Council under General Research Fund (Project no. 14225616), a grant from Hong Kong Innovation and Technology Commission under ITSP Tier 2 Fund (Project no. ITS/426/17FP), and a grant from Hong Kong Research Grants Council (Project no. PolyU 152035/17E).

\bibliographystyle{aaai}
\bibliography{reference}

\end{document}